\documentclass[10pt, a4paper, twocolumn]{article}
\usepackage[utf8]{inputenc}
\usepackage{amsmath}
\usepackage{graphicx}
\usepackage{geometry}
\usepackage{cite}          
\usepackage{hyperref}
\usepackage{float}
\usepackage{url}
\usepackage{setspace}
\usepackage{booktabs}      
\usepackage{titlesec}      
\usepackage{caption}       

\titlespacing*{\section}{0pt}{*2}{*1}
\titlespacing*{\subsection}{0pt}{*1.5}{*0.5}

\title{\textbf{A Real-Time, Vision-Based System for Badminton Smash Speed Estimation on Mobile Devices}}
\author{
    Diwen Huang \\
    \textit{Independent Researcher} \\
    \url{diwenhuang.ca} \\
    \texttt{diwenhuang.research@gmail.com}
}
\date{\today}

\begin{document}
\setstretch{1.1}

\twocolumn[
  \begin{@twocolumnfalse}
    \maketitle
    \begin{abstract}
        \noindent Performance metrics in sports, such as shot speed and angle, provide crucial feedback for athlete development. However, the technology to capture these metrics has historically been expensive, complex, and largely inaccessible to amateur and recreational players. This paper addresses this gap in the context of badminton, one of the world's most popular sports, by introducing a novel, cost-effective, and user-friendly system for measuring smash speed using ubiquitous smartphone technology. Our approach leverages a custom-trained YOLOv5 model for shuttlecock detection, combined with a Kalman filter for robust trajectory tracking. By implementing a video-based kinematic speed estimation method with spatiotemporal scaling, the system automatically calculates the shuttlecock's velocity from a standard video recording. The entire process is packaged into an intuitive mobile application, democratizing access to high-level performance analytics and empowering players at all levels to analyze and improve their game.
        \vspace{2em}
    \end{abstract}
  \end{@twocolumnfalse}
]


\section{Introduction}
The proliferation of data analytics has revolutionized modern sports, offering athletes and coaches unprecedented insights into performance. Key metrics such as shot speed have become fundamental for training, strategy, and talent identification. Despite these advancements, access to the requisite technology remains a significant barrier for those outside professional circuits, primarily due to prohibitive costs and operational complexity. While several efforts have sought to democratize sports technology, these innovations have not been adequately extended to badminton.

Badminton is a global sport, with participation levels that rank it among the top two most popular sports worldwide \cite{clement2004badminton}. Nevertheless, consumer-grade products for performance analysis are conspicuously scarce. The primary available tool, the radar gun, is not only exorbitant but also cumbersome to configure and operate. Furthermore, standard radar guns often struggle to consistently detect the badminton shuttlecock due to its small size and low radar cross-section, which impedes functionality \cite{edrich2012limitations}.

This paper presents a novel solution that circumvents these barriers. We have developed a system that is free, user-friendly, and universally accessible to anyone with a modern smartphone. By harnessing the power of computer vision and sophisticated tracking algorithms, our application provides a reliable method for measuring the pinnacle metric of badminton performance: the smash speed.

\section{Background and Related Work}
Arguably the most celebrated and crucial metric in badminton is the speed of the smash. Existing solutions for its measurement vary drastically in accessibility and accuracy.

\subsection{Professional Tournament Technology}
In elite international tournaments, systems like Hawk-Eye are employed to provide instantaneous and highly accurate data. This technology utilizes multiple high-speed cameras positioned around the court to triangulate the shuttlecock's position in 3D space. While representing the gold standard in accuracy, this solution is confined to professional venues and is entirely unachievable for the vast majority of players.

\subsection{Radar Guns}
The radar gun is the most common alternative. It measures speed by emitting radio waves and analyzing the Doppler shift in the reflected waves from a moving object. However, high-precision sports radar guns cost thousands of dollars and require careful positioning to yield accurate results. Cheaper consumer models often fail to reliably detect the shuttlecock, as its non-metallic feather or nylon construction provides a poor reflective surface for radar waves.

\subsection{Video-Based Analysis}
A more accessible but laborious method involves recording a video and manually tracking the shuttlecock frame by frame using software. After identifying the shuttlecock's pixel coordinates in consecutive frames, one can apply kinematic equations and a known reference distance to calculate its real-world speed. This process is exceedingly technical, time-consuming, and prone to human error, making it impractical for regular use.

Beyond 2D speed measurement, other research has focused on the more complex challenge of full 3D trajectory reconstruction from a single camera view, using geometric constraints and physics-based models to estimate the shuttlecock's flight path in three-dimensional space \cite{lee2019trajectory, shen2018reconstruction}. Our work, however, prioritizes accessibility and immediate feedback by focusing on the peak 2D velocity, which is the most critical metric for player training.

Our proposed solution directly addresses the shortcomings of these methods. It does not require professional status, thousands of dollars in equipment, or hours of tedious manual analysis. Instead, it offers an automated, on-demand tool for any player with a phone.

\begin{figure*}[htbp]
    \centering
    \includegraphics[width=\textwidth]{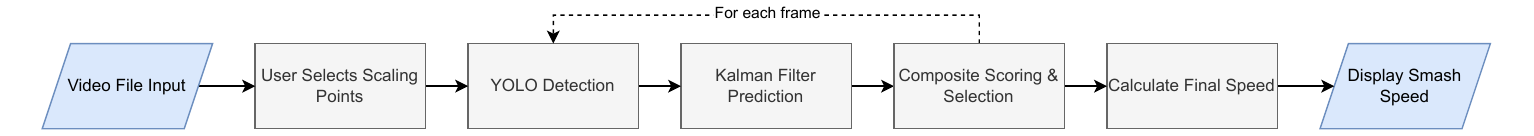}
    \caption{The end-to-end system pipeline. The user uploads a video, which is processed frame-by-frame by the YOLOv5 model. Detections are filtered and smoothed by the Kalman filter, and the resulting trajectory is used to calculate the peak smash speed.}
    \label{fig:pipeline}
\end{figure*}

\section{Methodology}
Our system provides an end-to-end pipeline for automated speed calculation using video-based kinematic speed estimation (see Figure \ref{fig:pipeline}). The core of our method involves spatiotemporal scaling and robust shuttlecock tracking via a custom-trained computer vision model, further refined with a state-estimation filter.

\subsection{Data Collection and Model Training}
\begin{figure}[htbp]
    \centering
    \includegraphics[width=0.48\columnwidth]{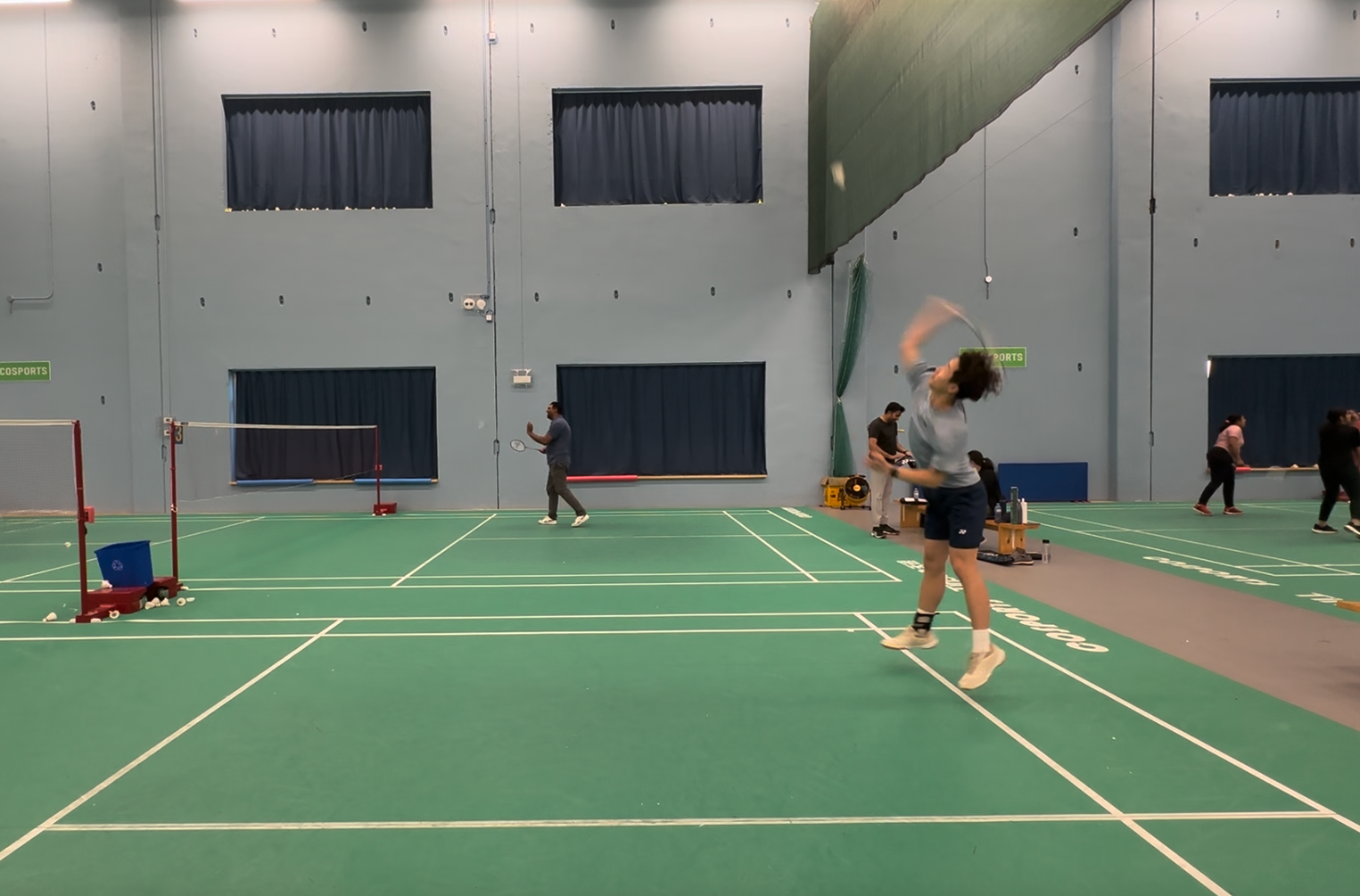}
    \caption*{(a) Correct Perpendicular View}
    \vspace{1em}
    \includegraphics[width=0.48\columnwidth]{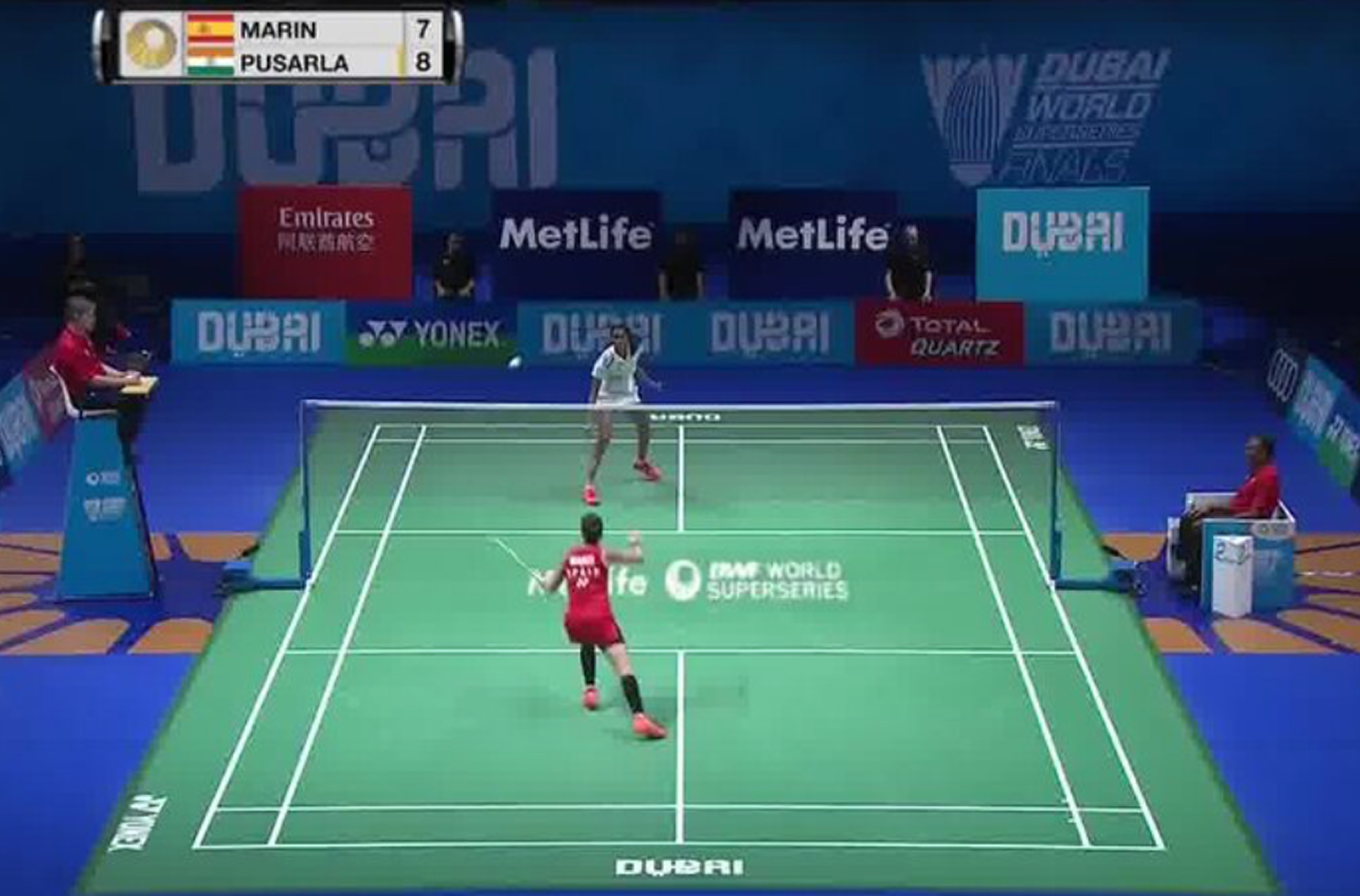}
    \caption*{(b) Unsuitable Broadcast View}
    \caption{Comparison of camera views. (a) The required perpendicular view for our 2D analysis. (b) A typical broadcast view, unsuitable due to perspective distortion.}
    \label{fig:camera_views}
\end{figure}

Our methodology relies on a strict two-dimensional analysis, which requires the camera to be positioned perpendicular to the shuttlecock's primary plane of motion, as shown in Figure \ref{fig:camera_views}(a). This orthogonal viewpoint effectively eliminates the z-axis (depth) from calculations, simplifying the problem to x and y coordinates. Consequently, established public datasets for ball tracking, such as TrackNetV2 \cite{sun2020tracknetv2}, which are typically captured from broadcast angles (Figure \ref{fig:camera_views}(b)), were unsuitable for our specific geometric constraints.

To address this, we first curated and annotated a custom dataset of 15,000 images, all captured from the required perpendicular perspective. This ensures the model is trained on data that mirrors the intended use case. We then used this dataset to train a YOLOv5 object detection model \cite{yolov5}. YOLOv5 was chosen for its excellent balance of speed and accuracy in real-world applications and its proven efficacy when trained on relatively small, custom datasets.

You Only Look Once (YOLO) is a family of single-pass object detection models. Unlike two-stage detectors, YOLO treats object detection as a single regression problem, directly from image pixels to bounding box coordinates and class probabilities \cite{redmon2016you}. Upon completion of training, the model achieved a \textbf{precision of 93\%}, a \textbf{recall of 87\%}, and a \textbf{mean Average Precision at an IoU threshold of 0.5 (mAP@0.5) of 91\%}, demonstrating a high of accuracy in identifying the shuttlecock.

\subsection{Automated Speed Calculation Program}
The program, developed in Python, orchestrates the entire analysis process.

\subsubsection{Spatiotemporal Scaling}
The user initiates the process by uploading a video and establishing a real-world scale factor. To correct for perspective distortion, the reference line for scaling must align with the player's depth on the court. The user selects two points on the court line (e.g., the rear doubles service line or the front service line) that is parallel to the camera's viewing plane and at the same distance from the camera as the player's point of impact. This ensures the shuttlecock and the reference line are in the same plane of motion, making the scale factor consistent. With a known real-world distance, $d_{real}$, the scale factor $S_f$ is computed from the pixel distance $d_{pixel}$:
$$S_f = \frac{d_{real}}{d_{pixel}} \quad [\text{meters/pixel}]$$

\subsubsection{Shuttlecock Detection and Trajectory Filtering}
The trained YOLOv5 model runs inference on every frame. To maximize recall, we use a low confidence threshold of 0.1 and an Intersection over Union (IoU) threshold of 0.45. This initial detection phase often produces a noisy trajectory. To refine this, we apply a multi-stage filtering process.

\paragraph{Heuristic Bounding.}
We first discard detections that are kinematically implausible. Any detection implying a speed below 5 km/h is rejected as a likely static false positive. Similarly, any detection implying a speed over 375 km/h is rejected to favor precision over recall for extreme outliers.

\paragraph{Kalman Filter Smoothing.}
To incorporate temporal context, we employ a linear Kalman filter \cite{kalman1960new, singh2018end}, which assumes a constant velocity motion model. This approach is a common and effective strategy for short-term predictions between frames and follows the tracking-by-detection paradigm, similar to algorithms like Simple Online and Realtime Tracking (SORT) \cite{bewley2016sort}. The filter predicts the shuttlecock's next state based on its current one and updates this prediction using the subsequent YOLO measurement.

\paragraph{Composite Scoring.}
For each plausible YOLO box, a composite score fuses the YOLO detection confidence with the box's proximity to the Kalman filter's prediction. To ensure the scoring is independent of video resolution, the proximity score, $P_{Kalman}$, is normalized. It is calculated based on the Euclidean distance, $d$, between the detection and the prediction, normalized by a factor of one-quarter of the frame's width, $W_{frame}$:
$$P_{Kalman} = \max\left(0, 1 - \frac{d}{W_{frame} / 4}\right)$$
This proximity score is then combined in a weighted sum with the YOLO confidence, $C_{YOLO}$, to produce the final composite score, $S_{comp}$. The weights are set empirically to prioritize temporal smoothness over raw detection confidence.
$$S_{comp} = (0.3 \cdot C_{YOLO}) + (0.7 \cdot P_{Kalman})$$
The detection with the highest composite score that passes the heuristic validation is chosen as the final position for the frame.

\subsubsection{User Verification and Speed Calculation}
The user can verify and correct the final trajectory. Speed is then calculated between each consecutive frame pair $(i, i+1)$. To enhance accuracy, the final speed is calculated not from the bounding box's geometric center, but from a motion-compensated point on the shuttlecock's leading edge. This method provides a more robust measurement, especially in frames with significant motion blur where the object appears elongated. With frame time $t_{frame} = 1/\text{FPS}$, the speed in km/h is:
$$v_{i \rightarrow i+1} = \frac{\sqrt{(x_{i+1}-x_i)^2 + (y_{i+1}-y_i)^2}}{t_{frame}} \times S_f \times 3.6$$
The peak speed is reported as the final result.

\subsection{Mobile Application Development}
To ensure accessibility, the system was packaged into user-friendly applications for both iOS (Swift) and Android (React Native).

\section{Experimental Validation}
\subsection{Setup}
The experiment was conducted on a standard badminton court. The equipment included an iPhone 16 recording at 30 FPS, a Bushnell Speedster III radar gun, and standard tournament-grade shuttlecocks. The choice to test at 30 FPS was intentional, as it represents a baseline for modern smartphones, ensuring the system's viability for the average user, not just those with high-end devices capable of 60 or 120 FPS. The radar gun was positioned behind the net, facing the player, serving as the ground truth. The phone was placed on a tripod perpendicular to the smash trajectory.

\subsection{Procedure}
A skilled player executed a series of 20 smashes. For each smash, the speed was recorded simultaneously by our application and the radar gun.

\subsection{Results}
The data collected revealed a significant and systematic discrepancy between the two systems when measuring peak velocity. Our vision-based application consistently recorded substantially higher speeds than the radar gun, as shown in Table \ref{tab:results}. The Mean Absolute Error (MAE) between the two methods for peak speed was 66.41 km/h, and the Root Mean Squared Error (RMSE) was 74.68 km/h. This large difference does not indicate an error in our system; rather, it highlights a fundamental limitation of the radar gun in this context.

\begin{figure}[htbp]
    \centering
    \includegraphics[width=\columnwidth]{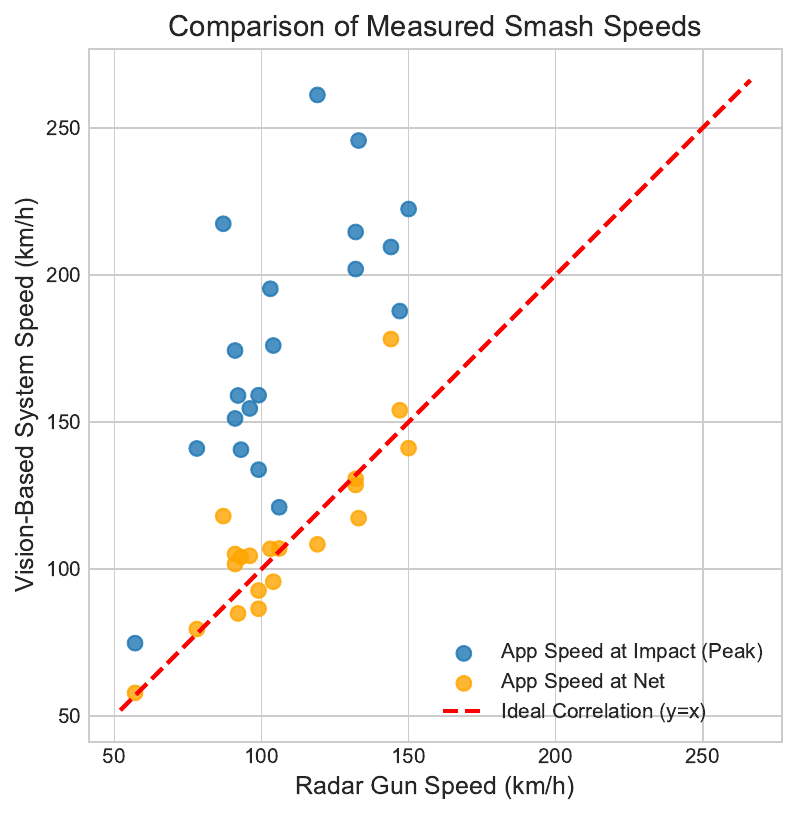}
    \caption{Scatter plot of speeds measured by our system versus the radar gun. The consistent position of points above the y=x line demonstrates the systematic difference in measurement, with our system capturing the higher peak velocity at impact.}
    \label{fig:scatterplot}
\end{figure}

This discrepancy is explained by the extreme deceleration of the shuttlecock due to aerodynamic drag \cite{cohen2015aerodynamics}. Our application measures the peak speed immediately after impact. The radar gun, however, requires the shuttlecock to travel a certain distance before it can lock on, by which time significant deceleration has already occurred. The scatter plot in Figure \ref{fig:scatterplot} visualizes this, showing a positive correlation but with all data points for peak speed far above the y=x line, confirming our system is measuring a different, earlier, and higher velocity.

To further validate this hypothesis, we used our application to measure the shuttlecock's speed at the approximate location where the radar gun would take its reading (near the net). As shown in Table \ref{tab:results} and Figure \ref{fig:scatterplot}, these "at-net" measurements from our system align remarkably well with the radar gun's readings. This confirms that our system is calibrated correctly and is accurately capturing the shuttlecock's velocity throughout its trajectory. The key finding is that our system measures the true peak velocity at impact, a metric the radar gun is physically incapable of capturing.

\begin{table*}[!htbp]
    \centering
    \caption{Comparison of Measured Smash Speeds (km/h) for all 20 Trials}
    \label{tab:results}
    \begin{tabular}{cccc}
        \toprule
        \textbf{Trial} & \textbf{Radar Gun} & \textbf{Our System (Peak Speed)} & \textbf{Our System (Speed at Net)} \\
        \midrule
        1 & 132.0 & 214.6 & 128.6 \\
        2 & 133.0 & 245.7 & 117.3 \\
        3 & 119.0 & 261.2 & 108.4 \\
        4 & 147.0 & 187.7 & 154.0 \\
        5 & 144.0 & 209.5 & 178.2 \\
        6 & 93.0 & 140.6 & 104.1 \\
        7 & 96.0 & 154.6 & 104.5 \\
        8 & 99.0 & 159.1 & 92.7 \\
        9 & 78.0 & 141.0 & 79.6 \\
        10 & 99.0 & 133.8 & 86.5 \\
        11 & 92.0 & 159.0 & 84.9 \\
        12 & 91.0 & 174.3 & 105.1 \\
        13 & 104.0 & 176.0 & 95.7 \\
        14 & 150.0 & 222.4 & 141.1 \\
        15 & 87.0 & 217.4 & 118.0 \\
        16 & 132.0 & 202.0 & 130.7 \\
        17 & 91.0 & 151.2 & 101.7 \\
        18 & 57.0 & 74.8 & 57.9 \\
        19 & 106.0 & 121.0 & 107.0 \\
        20 & 103.0 & 195.3 & 106.9 \\
        \bottomrule
    \end{tabular}
\end{table*}

\section{Conclusion}
This paper introduced a novel, vision-based system for accurately measuring badminton smash speeds using only a smartphone. The experimental validation demonstrated that our system provides a more relevant and accurate measurement of peak smash velocity than consumer-grade radar guns. By capturing the speed at the moment of impact, our system avoids the significant underestimation caused by the shuttlecock's rapid deceleration, a factor that fundamentally limits the utility of radar-based methods.

The key finding of this work is not just that a smartphone can measure smash speed, but that it can do so more effectively for training purposes by capturing the true peak velocity. By confirming that our system's speed readings near the net align with the radar gun's readings, we have validated its accuracy across the trajectory and proven its superiority in measuring the crucial metric of post-impact speed. The primary implication is the democratization of superior sports analytics, empowering players with feedback that was previously either inaccessible or inaccurate.

\subsection{Limitations}
Despite its strong performance, the system has several limitations. Its accuracy is contingent on a stable video recording and correct user input during the initial scaling phase. Furthermore, the strict 2D planar assumption means the system cannot accurately measure the speed of cross-court smashes. Any movement along the z-axis (towards or away from the camera) is not captured, leading to an underestimation of the true velocity for angled shots. However, this limitation is considered minor for the tool's primary use case, which assumes that players are practicing straight smashes where lateral deviation is minimal. The performance of the YOLOv5 model can also be affected by challenging lighting conditions, which may reduce detection accuracy. Finally, the validation was conducted with a limited sample size of 20 smashes and with smashes intentionally kept below maximum power due to the radar gun's limitations, and a larger, more varied dataset could provide more robust statistical confirmation of the system's accuracy.

\subsection{Future Work}
Future work will focus on addressing these limitations and expanding the system's capabilities. A key area for development is the expansion of analytics to include full 3D trajectory reconstruction. We plan to explore state-of-the-art monocular 3D tracking models, such as MonoTrack \cite{liu2022monotrackshuttle}, and train similar models on badminton-specific datasets like TrackNetV2 \cite{sun2020tracknetv2} to calculate not just speed, but also launch angle and trajectory arc from a single, uncalibrated video feed.


\bibliographystyle{IEEEtran}
\bibliography{references}

\end{document}